\documentclass{article}

\usepackage{times}
\usepackage{graphicx} % more modern
\usepackage[tight]{subfigure}

\usepackage{bm}
\usepackage{amsmath}
\usepackage{amsfonts}
\usepackage{mathtools}

\usepackage{natbib}

\usepackage{algorithm}
\usepackage{algorithmic}

\usepackage{hyperref}

\usepackage[accepted]{icml2017_arxiv}

\icmltitlerunning{Learning Texture Manifolds}

\usepackage{cancel}
\usepackage{xargs}

\newcommand\blfootnote[1]{%
  \begingroup
  \renewcommand\thefootnote{}\footnote{#1}%
  \addtocounter{footnote}{-1}%
  \endgroup
}

\begin{document} 

\twocolumn[
%\vspace*{-2cm}
\icmltitle{Learning Texture Manifolds with the Periodic Spatial GAN}

\icmlauthor{Urs Bergmann*}{urs.bergmann@zalando.de}
\icmlauthor{Nikolay Jetchev*}{nikolay.jetchev@zalando.de}
\icmlauthor{Roland Vollgraf}{roland.vollgraf@zalando.de}
\icmladdress{Zalando Research, Berlin\\[-0.1em] \textit{\scriptsize *authors contributed equally}\\}

\icmlkeywords{generative model, GAN, texture, adversarial, machine learning, ICML}

\vskip 0.3in
]

\begin{abstract} 
This paper introduces a novel approach to texture synthesis based on generative adversarial networks (GAN)~\cite{Goodfellow14}. We extend the structure of the input noise distribution by constructing tensors with different types of dimensions. We call this technique Periodic Spatial GAN (PSGAN). 

The PSGAN has several novel abilities which surpass the current state of the art in texture synthesis. 
First, we can learn multiple textures from datasets of one or more complex large images. Second, we show that the image generation with PSGANs has properties of a texture manifold: we can smoothly interpolate between samples in the structured noise space and generate novel samples, which lie perceptually between the textures of the original dataset.
In addition, we can also accurately learn periodical textures. 
We make multiple experiments which show that PSGANs can flexibly handle diverse texture and image data sources.
Our method is highly scalable and it can generate output images of arbitrary large size.
\end{abstract} 

\vspace{-0.4cm}

\section{Introduction}
\subsection{Textures and Texture Synthesis}

\begin{figure*}[t]
\centering
\includegraphics[height=10.5cm]{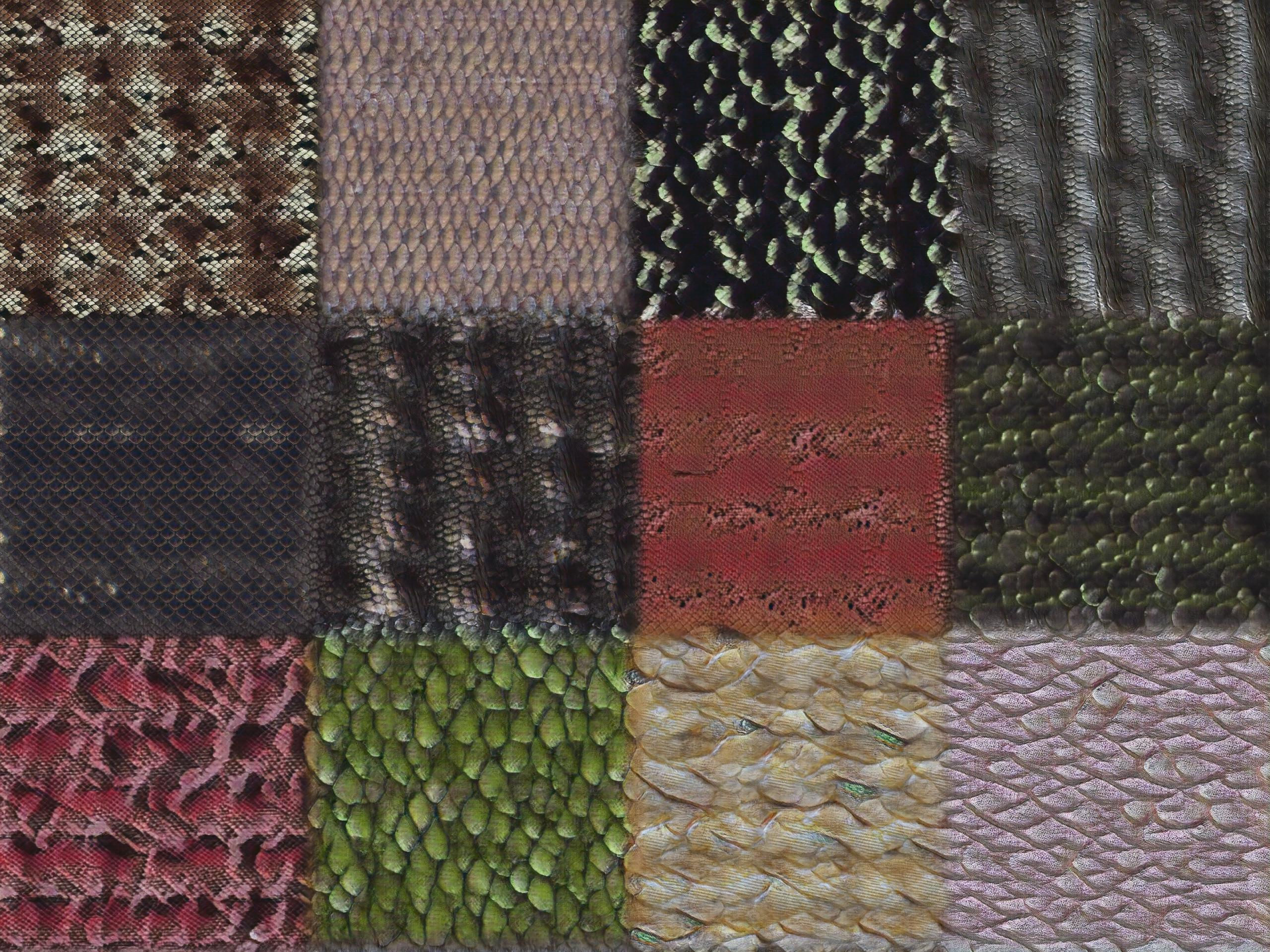} 
\caption{PSGANs can extract textures from complex datasets of natural images, here the Oxford Describable Textures Dataset~\cite{cimpoi14describing} - category ``scaly". The image shows a quilt of 3x4 different tiles, each containing a novel synthesized texture, not originally in the dataset. Methodologically, the image is created by setting the global dimensions of the $Z$ tensor for local regions of $20 \times 20$ spatial dimensions to be identical, resulting in an image of a total size 1920x2560 in pixels.}\label{fig_scaly}
\end{figure*}
Textures are important perceptual elements, both in the real world and in the visual arts. Many textures have random noise characteristics, formally defined as stationary, ergodic, stochastic processes ~\cite{DCC2013}.
There are many natural image examples with such properties, e.g.\ rice randomly spread on the ground. However, more complex textures also exist in nature, e.g.\ those that exhibit periodicity like a honeycomb or fish scales.

The goal of texture synthesis is to learn from a given example image a generating process, which allows to create many images with similar properties.
Classical texture synthesis methods include instance based approaches~\cite{EfrosP,EfrosQ}, where pixels or patches of the source image are resampled and copied next to similar image regions, so that a seamless bigger texture image is obtained. Such methods have good visual quality and can deal with periodic images, but have a high runtime complexity when generating big images. In addition, since they do not learn an explicit model of images but just copy patches from the original pixels, they cannot be used to generate novel textures from multiple examples.

Parametric methods define an explicit model of a ``good" texture by specifying some statistical properties; new texture images that are optimal w.r.t.\ the specified criteria are synthesized by optimization. The method of~\cite{Portilla:2000} yields good results in creating various textures, including periodic ones (the parametric statistics include phase variables of pre-specified periodicity).
However, the run-time complexity is high, even for small output images.
The authors also tried blending of textures, but the results were not satisfactory: patch-wise  mixtures were obtained, rather than a new homogeneous texture that perceptually interpolates the originals. 

More recently, deep learning methods were shown to be a powerful, fast and data-driven, parametric approach to texture synthesis.
The work of~\cite{Gatys2015b} is a milestone: they showed that filters from a discriminatively trained deep neural network can be used as effective parametric image descriptors. Texture synthesis is modeled as an optimization problem.~\cite{GatysEB15a} also showed the interesting application of painting a target content photo in the style of a given input image: ``neural art style transfer". Related works speed-up texture synthesis and style transfer by approximating the optimization process by feed-forward convolutional networks~\cite{ulyanov16texture,Johnson2016Perceptual}. 

However, the choice of descriptor in all of these related works -- the Gram matrix of learned filters -- is a specific prior on the learnable textures for the method. It generalizes to many, but not all textures -- e.g.\ periodic textures are reproduced inaccurately. 
Another limitation is that texture synthesis is performed from a single example image only, lacking the ability to represent and morph textures defined by several different images. In a related work,~\cite{DumoulinSK16} explored the blending of multiple styles by parametrically mixing their statistical descriptors. The results are interesting in terms of image stylization, but the synthesis of novel blended textures has not been shown.

\subsection{GANs}
Purely data driven generative models are an alternative deep learning approach to texture synthesis. 
Introduced in~\cite{Goodfellow14}, generative adversarial networks (GAN) train a model $G$ that learns a data distribution from example data, and a discriminator $D$ that attempts to distinguish generated from training data.
The GAN architecture was further improved~\cite{RadfordMC15} by using deep convolutional layers with (fractional) stride.
GANs have successfully created ``natural" images of great perceptual quality that can fool even human observers. However, pixel resolution is usually low, and the output image size is pre-specified and fixed at training time. 

For the texture synthesis use case, fully convolutional layers, which can scale to any image size, are advantageous.~\cite{LiW16b} presented an interesting architecture, that combines ideas from GANs and the pre-trained descriptor of~\cite{Gatys2015b} in order to generate small patches with the statistics of layer activations from the VGG network. This method allows fast texture synthesis and style transfer.

Spatial GAN (SGAN)~\cite{SGAN2016} applied for the first time fully unsupervised GANs for texture synthesis. SGANs had properties like good scalability w.r.t.\ speed and memory, and showed excellent results on certain texture classes, surpassing the results of~\cite{Gatys2015b}. However, some classes of textures cannot be handled, and no plausible texture morphing is possible.

The current contribution, PSGAN, makes a great step forward with respect to the types of images a neural texture synthesis method can create -- both periodic and non-periodic images are learned in an unsupervised way from single images or large datasets of images. Afterwards, flexible sampling in the noise space allows to create novel textures of potentially infinite output size, and smoothly transition between them. Figure \ref{fig_scaly} shows a few example textures generated with a PSGAN.
In the next section we describe in detail the architecture of the PSGAN, and then proceed to illustrate its abilities with a number of experiments\blfootnote{Our source code is available at \url{https://github.com/zalandoresearch/psgan}}.

\section{Methods: Periodic GAN}
\label{sec:pgan_method}

\begin{figure*}[tb]
 \center{
  \includegraphics[width=0.75\textwidth]{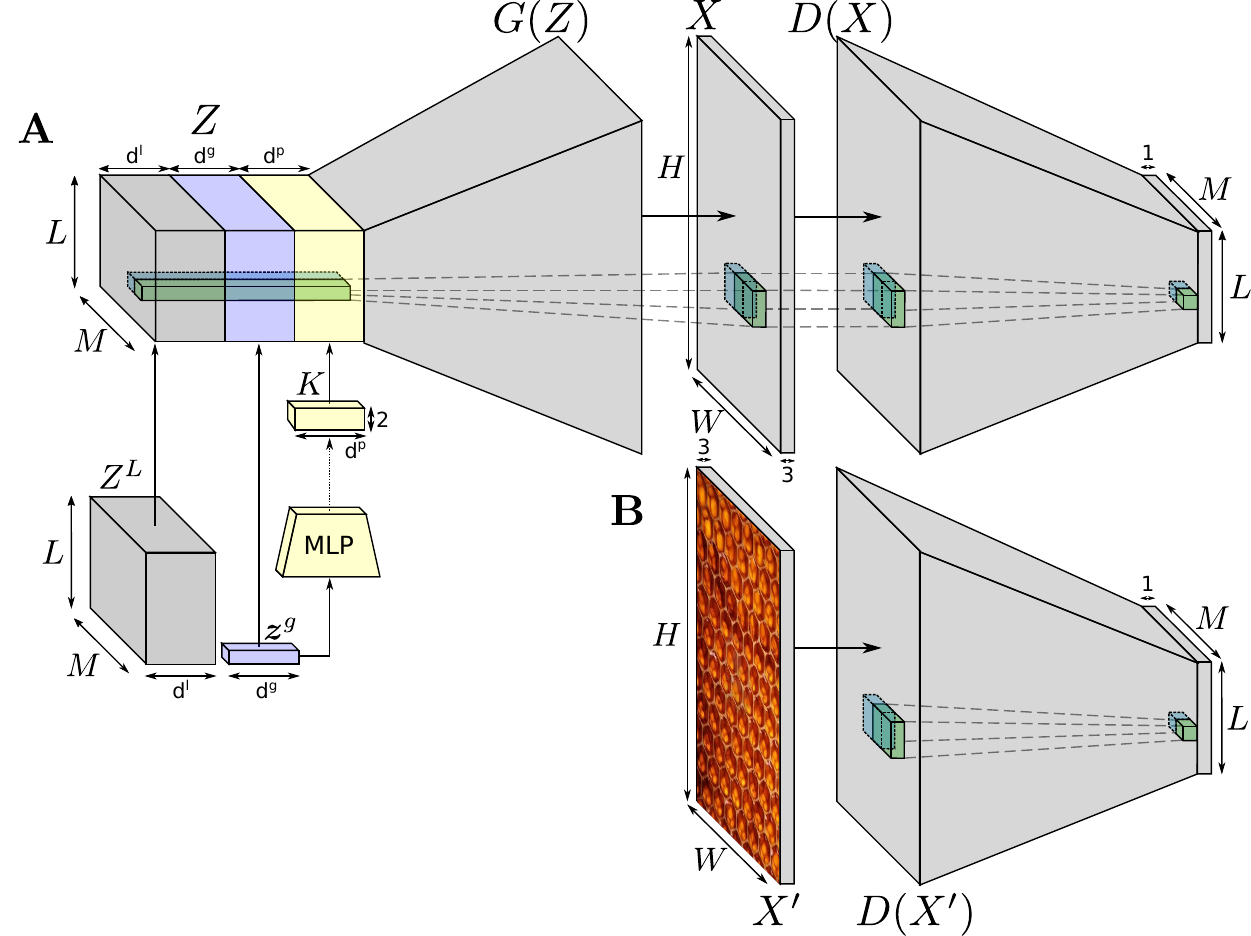} 
 }
 \caption{Illustration of the PSGAN model. \textbf{A} The fully convolutional generator network $G(Z)$ maps a spatial tensor $Z_{\lambda \mu i}$, $\lambda$ and $\mu$ being the spatial indices, to an input image $X$. Every subvector at a spatial location in $Z$, e.g.\ the blue or green columns in the Figure, map to a limited area in $X$. To alleviate the independence property of distant areas in $X$ we construct the $Z$ tensor out of three parts: a local part $Z^l$, a global part $Z^g$ and a periodic part $Z^p$ -- see text. As usual in GAN training, the discriminator gets either a generated image $X$ or, as in \textbf{B}, an image patch $X'$ from the real data distribution.}
  \label{fig:PSGAN}
\end{figure*}

In GANs, the generative model $G(\bm{z})$ maps a noise vector $\bm{z}$ to the input data space. As in SGANs ~\cite{SGAN2016}, we generalize the generator $G(Z)$ to map a noise tensor $Z \in \mathbb{R}^{L \times M \times d}$ to an image $X \in \mathbb{R}^{H \times W \times 3}$, see Figure~\ref{fig:PSGAN}. The first two dimensions, $L$ and $M$, are spatial dimensions, and are blown up by the generator $G(Z)$ to the respective input spatial dimensions $H>L$ and $W>M$. The final dimension of $Z$, $d$, is the channel dimension.

In analogy to the extension of the generator $G$, we extend the discriminator $D$ to map from an input image $X$ to a two-dimensional field of spatial size $L \times M$. 
Each position of the resulting discriminator $D_{\lambda \mu}(X), \; 1\leq \lambda \leq L \; \mathrm{and} \; 1\leq \mu \leq M$, responds only to a local part $X$, which we call $D_{\lambda \mu}$'s effective receptive field. The response of $D_{\lambda \mu}(X)$ represents the estimated probability that the respective part of $X$ is real instead of being generated by $G$. 

As the discriminator outputs a field, we extend the standard GAN cost function $V(D,G)$ to marginalize spatially:
\begin{align}
V(D,G) =& \frac{1}{LM} \sum_{\lambda=1}^L \sum_{\mu=1}^M \mathbb{E}_{Z \sim p_Z(Z)} \left[ \log \left( 1 - D_{\lambda\mu}(G(Z)) \right) \right]  \nonumber \\
  +& \frac{1}{LM} \sum_{\lambda=1}^{L} \sum_{\mu=1}^{M} \mathbb{E}_{X' \sim p_{\mathrm{data}}(X)} \left[ \log D_{\lambda\mu}(X') \right]
  \label{eq:pgan_vfunc}
\end{align}
This function is then minimized in $G$ and maximized in $D$, $ \min_G \max_D V(D,G)$. Maximizing the first line of eq.~\ref{eq:pgan_vfunc} in $D$ leads the discriminator to return values close to $0$ (i.e.\ ``fake") for generated images -- and, vice versa, minimization in $G$ aims at the discriminator taking large output values close to $1$ (i.e.\ ``real"). On the other hand, maximizing $D$ in the second line of eq.~\ref{eq:pgan_vfunc} anchors the discriminator on real data $X' \sim p_{\mathrm{data}}(X)$ to return values close to $1$. As we want the model to be able to learn from a single image, the input image data is augmented by selecting patches $X'$ from the image(s) at random positions. To speed-up convergence, in particular in the beginning of the learning process, we employ the standard GAN trick and substitute $\log(1-D(G(Z)))$ with $- \log(D(G(Z)))$~\cite{Goodfellow14}.

We base the design of the generator network $G$ and the discriminator network $D$ on the DCGAN model~\cite{RadfordMC15}. Empirically, choosing $G$ and $D$ to be symmetric in their architecture (i.e.\ depth and channel dimensions) turned out to stabilize the learning dynamics. In particular, we chose equal sizes for the image patches $X'$ and the generated data $X=G(Z)$. As a deviation from this symmetry rule, we found that removing batch normalization in the discriminator yields better results, especially on training with single images.

In contrast to the DCGAN architecture, our model contains exclusively convolutional layers. 
Due to the convolutional weight sharing, this allows that a network $G$ trained on small image patches $X$ can be rolled out to synthesize arbitrary large output images after training. Upon successful training, the sampled images then match the local image statistics of the training data.
Hence, the model implements a spatial stochastic process. Further, if components of $Z$ are sampled independently, the limited receptive fields of the generator $G$ imply that the generator implements a stationary, ergodic and strongly mixing stochastic process. This means that sampling of different textures is not possible -- this would require a non-ergodic process. 
For independent $Z$ sampling, learning from a set of textures results in the generation of textures combining elements of the whole set.
Another limitation of independent sampling is the impossibility to align far away regions in the generated image -- alignment violates translation invariance, stationarity and mixing. However, periodic textures depend on long-range correlations. 

To get rid of these limitations, we extend $Z$ to be composed of three distinct parts: a local independent part $Z^l$, a spatially global part $Z^g$, and a periodic part $Z^p$. Each part has the same spatial dimensions $L,M$, but may vary in their respective channel dimensions $d^l$, $d^g$, and $d^p$. Let $Z=[Z^l,Z^g,Z^p]$ be their concatenation with total channel dimension $d=d^l + d^g + d^p$. We proceed with a discussion on $Z$'s three parts.

\subsection{Local Dimensions}
Conceptually, the simplest approach is to sample each slice of $Z^l$ at position $\lambda$ and $\mu$, i.e $\bm{z}^l_{\lambda\mu} \in \mathbb{R}^{d^l}$, independently from the uniform distribution $p(\bm z)$,
where $\lambda, \mu \in \mathbb{N}$ with $1\leq \lambda \leq L$ and $1\leq \mu \leq M$. As each $\bm{z}^l_{\lambda\mu}$ affects a finite region in the image, we speak of local dimensions. Intuitively, local dimensions allow the generative process to produce spatial variance and diversity by sampling from its statistical model. 

\subsection{Global Dimensions}
For the global dimensions, a unique vector $\bm{z}^g$ of dimensionality $d^g$ is sampled from $p(\bm{z})$, which is then repeated along all $L \times M$ spatial dimensions of $Z^g$, or $Z^g_{\lambda \mu i}=\bm{z}^g_{i}$, where $1\leq \lambda \leq L$, $1\leq \mu \leq M$, and $1 \leq i \leq d^g$. Thus, $\bm{z}^g$ has global impact on the whole image, and  allows for the selection of the type of structure to be generated -- employing global dimensions, the generative stochastic process becomes non-ergodic. Consider the task of learning from two texture images: the generator then only needs to ``learn" a splitting of $\mathbb{R}^{d^g}$ in two half-spaces (e.g.\ by learning a hyperplane), where vectors $\bm{z}^g$ from each half-space generate samples in the style of one of the two textures.

Besides the scenario of learning from a set of texture images, combination with random patch selection from a larger image (see Section~\ref{sec:pgan_method}) is particularly interesting: here, the converged generator $G$ samples textures that are consistent with the local statistics of an image. Notably, the source image does not necessarily have to be a texture, but the method will extract a texture generating stochastic process from the image, nevertheless (see Figure~\ref{fig_rocinha}).

After learning, each vector $\bm{z}^g$ represents a texture from the manifold of learned textures of the PSGAN, where $\bm{z}^g$ corresponds to a generating stochastic process of a texture, not just a static image. 
For the purpose of image generation, $Z^g$ does not need to be composed of a single vector, but can be a smooth function in $\lambda$ and $\mu$.
As long as neighboring vectors in $Z^g$ don't vary too rapidly, the statistics of $Z^g$ is close to the statistics during training. Hence, smoothness in $Z^g$ implies a smooth texture change in $X$ (see Figure~\ref{scalemorph}).

\subsection{Spatially Periodic Dimensions}
The third part of $Z$, $Z^p$, contains spatial periodic functions, or plane waves in each channel $i$:
\begin{align}
 Z^p_{\lambda \mu i} = \zeta_{\lambda \mu i}(K) = \sin\left( \bm{k}_i^T \begin{pmatrix} \lambda \\ \mu \end{pmatrix} + \phi_i \right),
\end{align}
where $1\leq \lambda \leq L$, $1\leq \mu \leq M$, $1 \leq i \leq d^p$, and $K$ is a $2 \times d^p$ matrix which contains the wave vectors $\bm{k}_i$ as its column vectors. These vectors parametrize the direction and the number of radians per spatial $Z$ unit distance in the periodic channel $i$. $\phi_i$ is a random phase offset uniformly sampled from $[0,2\pi)$, and mimics the random positional extraction of patches from the real images.
Adding this periodic global tensor breaks translation invariance and stationarity of the generating process. However, it is still cyclostationary. 

While wave numbers $K$ could be set to a fixed basis, we note that a specific texture has associated wave vectors, i.e.\ different textures will have different axes of periodicities and scales. Hence, we make $K$ dependent on the global dimensions $\bm{z}^g$ through a multi-layer perceptron (MLP), when more than one texture is learned. When only one texture is learned, i.e.\ $d^g=0$, the wave numbers $K$ are direct parameters to the system. In Figure~\ref{fig:PSGAN}, we indicate this alternative dependency on $\bm{z}^g$ with a dotted arrow between the MLP and $K$.
All parameters of the MLP are learned end-to-end alongside the parameters of the generator $G$ and the discriminator $D$.

\section{Experiments}

\subsection{Experimental Setup}
We base our system on the DCGAN architecture~\cite{RadfordMC15} with a stride of $\frac{1}{2}$ for the generator and 2 for the discriminator. Local and global noise dimensions are sampled from a uniform distribution. As in DCGAN, filters have 64 channels at the highest spatial resolution, and are doubled after every layer, which halves the spatial resolution. E.g. the 4 layer architecture has $64-128-256$ channels between the noise input and output RGB image. Training was done with ADAM~\cite{KingmaB14} with the settings of~\cite{RadfordMC15} -- learning rate $0.0002$, minibatch size of 25. The typical image patch size was 160x160 pixels.
We usually used 5 layers in $G$ and $D$ (see Table~\ref{table}), kernels of size 5x5 with zero padding, and batch normalization. Such a generator upsamples the spatial noise by a factor of $\frac{H}{L}=\frac{W}{M}=32$ and has a receptive field size of 125. Receptive field and image patch size can both affect learning~\cite{SGAN2016}. 
On our hardware (Theano and Nvidia Tesla K80 GPU) we measured $0.006$ seconds for the generation of a 256x256 pixels image and $0.26$ seconds for a 2048x2048 pixels image.

The MLP for the spatially periodic dimensions has one hidden layer of dimensionality $d^h$:
\begin{equation}
K = \xi(\bm{z}^g) =   \begin{pmatrix} \left( W_1 f(W \bm{z}^g + \bm{b}) + \bm{b}_1 \right)^T \\
              		          \left( W_2 f(W \bm{z}^g + \bm{b}) + \bm{b}_2 \right)^T \end{pmatrix}, \nonumber
\end{equation}
where $f$ is the point-wise rectified-linear unit function, and we have $W \in \mathbb{R}^{d^h \times d^g}$, $\bm{b} \in \mathbb{R}^{d^h}$, $W_1$ and $W_2 \in \mathbb{R}^{d^p \times d^h}$, $\bm{b}_1$ and $\bm{b}_2 \in \mathbb{R}^{d^p}$. 
We used $d^h=60$ for the experiments.
All parameters are initialized from an independent random Gaussian distribution $\mathcal{N}(0,0.02)$, except $\bm{b}_1$ and $\bm{b}_2$, which have a non-zero mean $\mathcal{N}(\bm{c},0.02\bm{c})$. 
The constant vector $\bm{c} \in \mathbb{R}^{d^p}$ is chosen with entries spread in the interval $(0,\pi]$\footnote{Ideally, the wave numbers $K_{ji}$, with $j \in \{1,2\}$, should be within the valid interval between the negative and positive Nyquist wave numbers (here $[-\pi, \pi]$). However, wave numbers of single sinusoids are projected back into this interval. Hence, no constraint is necessary.}. 
For simplicity, we write $\varphi_{\lambda \mu i}(\bm{z}^g) \coloneqq \zeta_{\lambda \mu i}(\xi(\bm{z}^g)) = Z^p_{\lambda \mu i}$, or briefly $Z^p = \varphi(\bm{z}^g)$, to summarize the way the periodic dimensions arise from the global ones. Alternatively, for $Z^g$ not being composed of a single vector $\bm{z}^g$, we write for simplicity $Z^p = \varphi(Z^g)$ and understand this as $Z^p_{\lambda \mu i} =  \varphi_{\lambda \mu i}(Z^g_{\lambda \mu i}) = \zeta_{\lambda \mu i}(\xi(Z^g_{\lambda \mu}))$, where $Z^g_{\lambda \mu}$ denotes the vector slice in $Z^g$ along its last (i.e.\ $i$) dimension.

The following image sources were used for the experiments in this paper: 
the Oxford Describable Textures Dataset (DTD)~\cite{cimpoi14describing}, which is composed of various categories, each containing $120$ images; the Facades dataset~\cite{Tylecek13}, which contains 500 facades of different houses in Prague. Both datasets comprise objects of different scales and sizes.
We also used satellite images of Sydney from Google Maps. The P6 and Merrigum house are from Wikimedia Commons.

\subsection{Learning and Sampling Textures}
What are criteria for good texture synthesis? The way humans perceive a texture is not easily quantifiable with a statistic or metric. Still, one can qualitatively assess whether a texture synthesis method captures the right properties of a source image. In order to illustrate this, we will demonstrate how we can learn complex periodic images and texture manifolds, which allow texture blending.

\begin{table}
\centering
\begin{tabular}{|c|c|c|c|c|}
\hline 
image & $d^g$ & $d^l$ & $d^p$ & layer depth\\ 
\hline 
text, P6 & 0 & 10 & 2 & 4\\ 
\hline 
single honeycomb & 0 & 10 & 2 & 5 \\ 
\hline 
Merrigum & 10 & 30 & 2 &5 \\ 
\hline 
DTD & 40 & 20 & 4 &5 \\ 
\hline 
Facades & 40 & 20 & 6 &5 \\ 
\hline 
Sydney & 30 & 20 & 4 &5 \\ 
\hline 
\end{tabular} 
\caption{The dimension cardinality we used for different experiments. Note that when $d^g=0,d^p>0$ the MLP for wave number learning simplifies to just learning the bias values $\bm{b}_1,\bm{b}_2$.}\label{table}
\end{table}

\subsubsection{Periodic textures}
\begin{figure}[t]
\centering
\includegraphics[height=3.5cm]{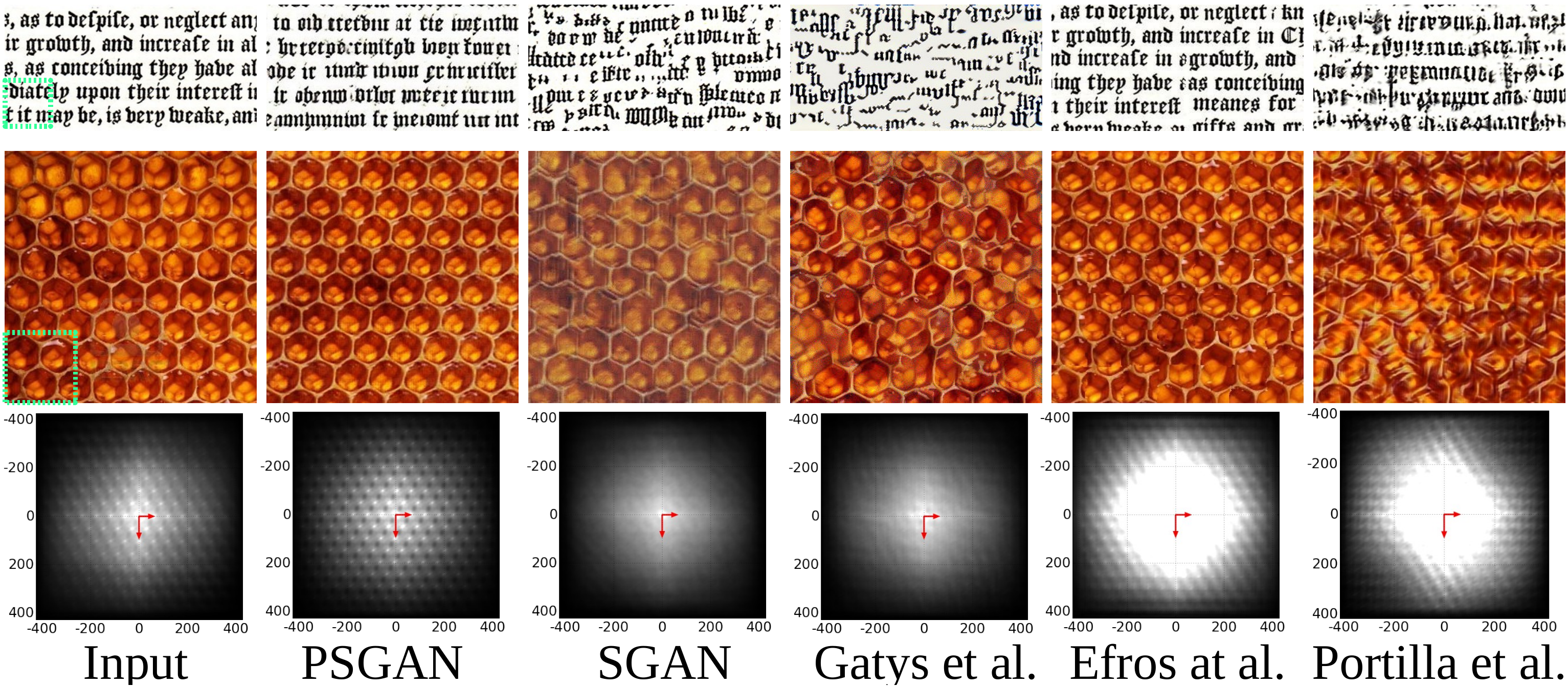}
\caption{
Comparing the results of 5 neural texture synthesis methods on 2 input images, text (168x336 pixels) in the top row and honeycomb (427x427 pixels) in the middle row. The green boxes show the receptive fields of the generator, 61x61 pixels for the text and 125x125 for the honeycomb example. In both cases, PSGAN best captures the underlying data periodicity. The autocorrelation plot of the honeycomb, shown in the bottom row, reveals periodicity as a grid of intensity peaks. The red arrows are the periodicity (inverse wave numbers) of the PSGAN, which neatly align with the autocorrelation  structure (best seen zoomed).}\label{fig_period1}
\end{figure}

First, we demonstrate learning a single periodic texture image.
Figure \ref{fig_period1} illustrates the results of PSGAN compared with SGAN~\cite{SGAN2016}, and the methods of~\cite{Gatys2015b, EfrosQ,Portilla:2000}. 
The text example in the top row has a periodic and stochastic dimension. The PSGAN learns  this and arranges ``text" in regular lines, while varying their content horizontally. The methods of~\cite{EfrosQ,Portilla:2000} also manage to do this. SGAN (equivalent to a PSGAN without periodic dimensions) and Gatys' method fail to capture the periodic structure.
 
The second row in Figure \ref{fig_period1} demonstrates learning a honeycomb texture -- a basic hexagonal pattern -- where our method captures both the underlying periodicity and the random coloring effects inside the cells. The method of \cite{EfrosQ} was inaccurate for that texture -- the borders between the copied patches (60x60 pixels large) were inaccurately aligned. The other 3 methods fail to produce a hexagonal structure even locally.
The last row of the figure shows the autocorrelation plots of the honeycomb textures, where the periodicity reveals itself as a regular grid superimposed onto the background, a feature only PSGAN is able to reproduce. 

While $d^p=2$ periodic dimensions are enough to learn the above patterns, we noticed that training convergence is faster when setting $d^p >2$. However, for $d^p > 2$ beating of sinusoids with close wave numbers can occur, which rarely happens also for $d^p \leq 2$ due to sub-Nyquist artefacts~\cite{amidror2015sub}, i.e.\ when the texture periodicity is close to an integer fractional of the Nyquist wavenumber.

\begin{figure}[tb]
\centering
\subfigure[\textbf{A} Honeycomb]{\includegraphics[height=0.92cm]{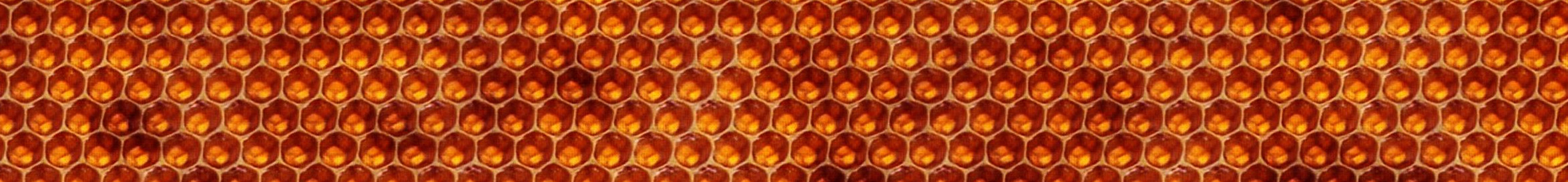}}
\vspace{-0.2cm}
\subfigure[\textbf{B} P6 wallpaper group]{\includegraphics[height=1.0cm]{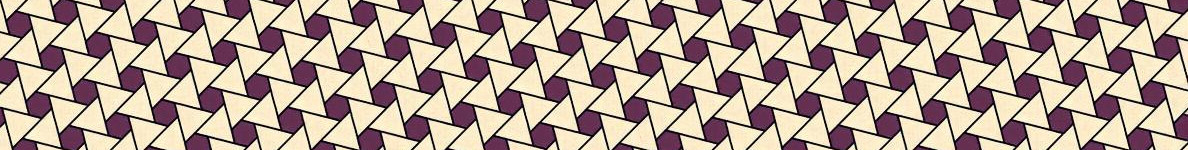}} 
\caption{Accurate wave number learning by PSGAN allows correct generation of periodic textures, even for very large images. \textbf{A} the honeycomb can be repeated in large images (300x2400 pixels), with no aliasing. \textbf{B} the non-orthogonal bases of the periodicities and complicated symmetries and rotations present in the P6 pattern are faithfully reproduced.}\label{fig_period2}
\end{figure}

Figure \ref{fig_period2} shows a larger slice of the learned periodic textures. In particular, Figure~\ref{fig_period2}\textbf{B} shows that learning works for more complex patterns, here a pattern with a P6 wallpaper group symmetry\footnote{\url{en.wikipedia.org/wiki/Wallpaper_group}. Note that only translational symmetries are represented in PSGANs, no rotation and reflection symmetries.}, with non-orthogonal symmetry axes.


\subsubsection{Texture manifolds}\label{sec_rocinha}
Next, we extract multiple textures from a single large image, or a set of images.
The chosen images (e.g.\ landscape photography or satellite images) have a global structure, but also exhibit characteristics of many textures in a single image (e.g.\ various vegetation and houses).
The structured PSGAN generator noise with global dimensions allows to extract textures, corresponding to different image regions. 

In order to visualize the texture diversity of a model, we define a \textit{quilt} array that can generate different textures from a trained PSGAN model by setting rectangular spatial regions (tiles) $Z^g_{\lambda:\lambda+\Delta,\mu:\mu+\Delta}$ of size $\Delta \times \Delta$ to the same vector, randomly sampled from the prior. Since the generator is a convolutional network with receptive fields over several spatial elements of $Z^g$, the borders between tiles look partially aligned. For example, in Figure \ref{fig_scaly} the borders of the tiles have scaly elements across them, rather than being sharply separated (as the input $Z^g$ per construction). 

\begin{figure*}[tb]
\centering
\subfigure[Input]{\includegraphics[height=4.1cm]{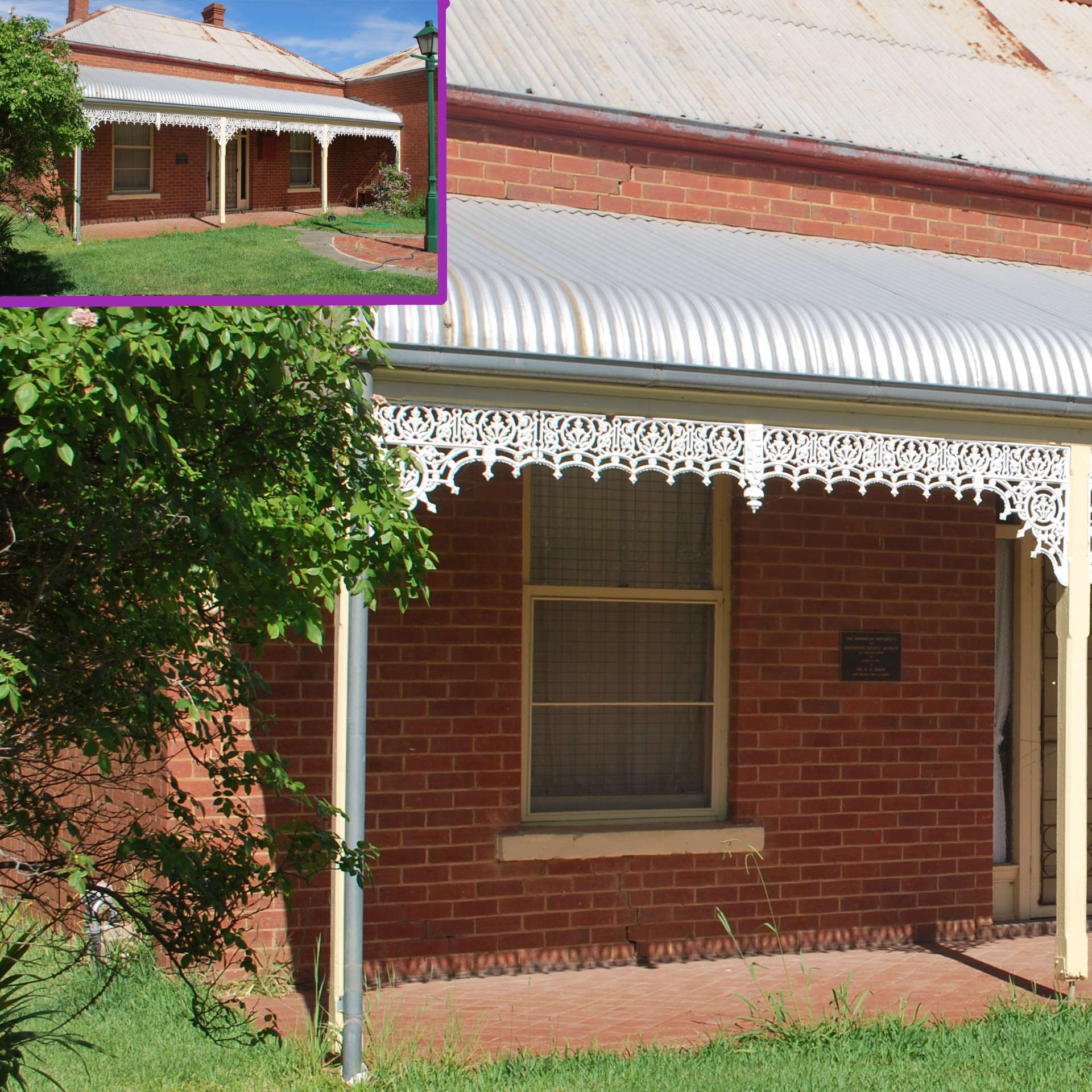} }
\subfigure[PSGAN]{\includegraphics[height=4.1cm]{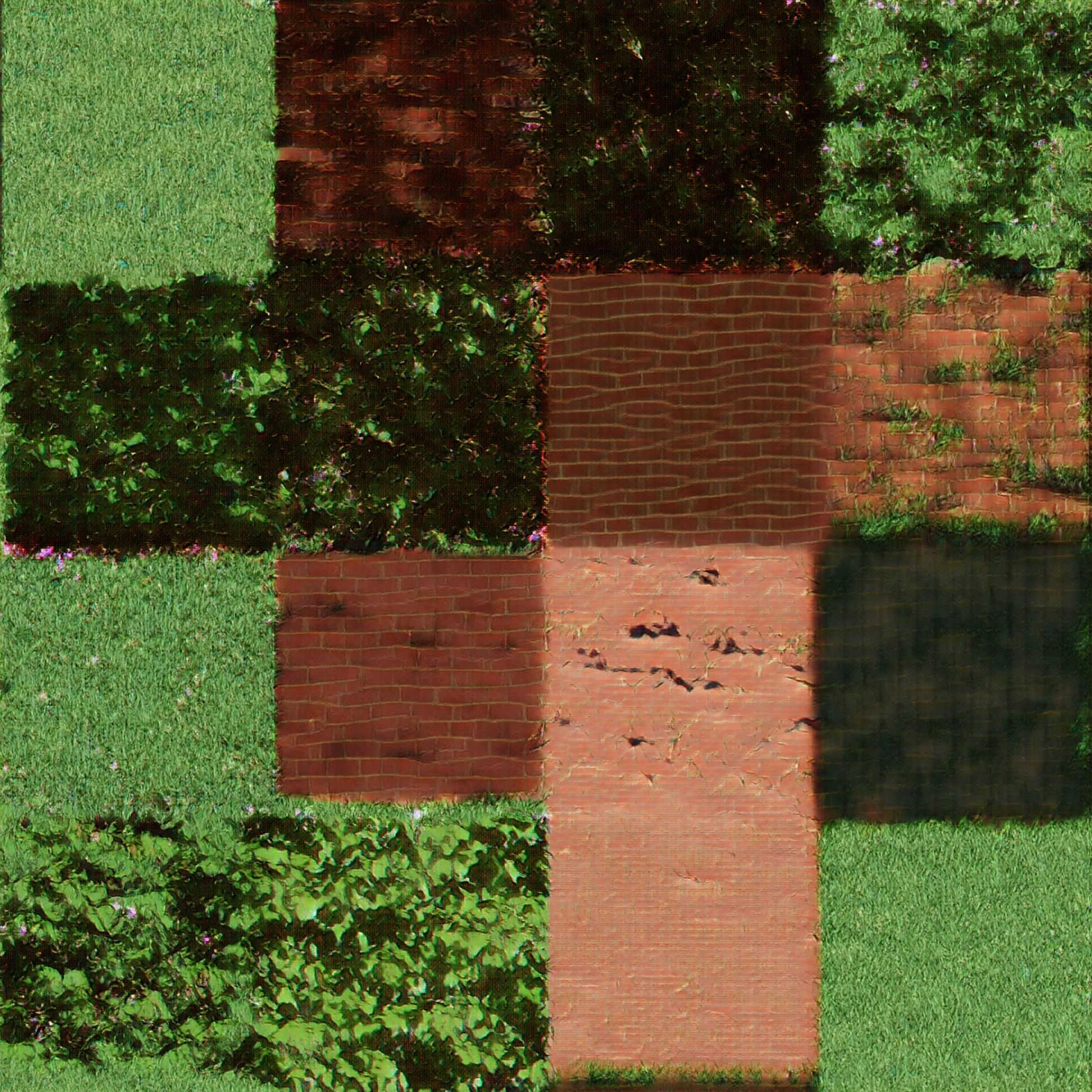}} 
\subfigure[SGAN]{\includegraphics[height=4.1cm]{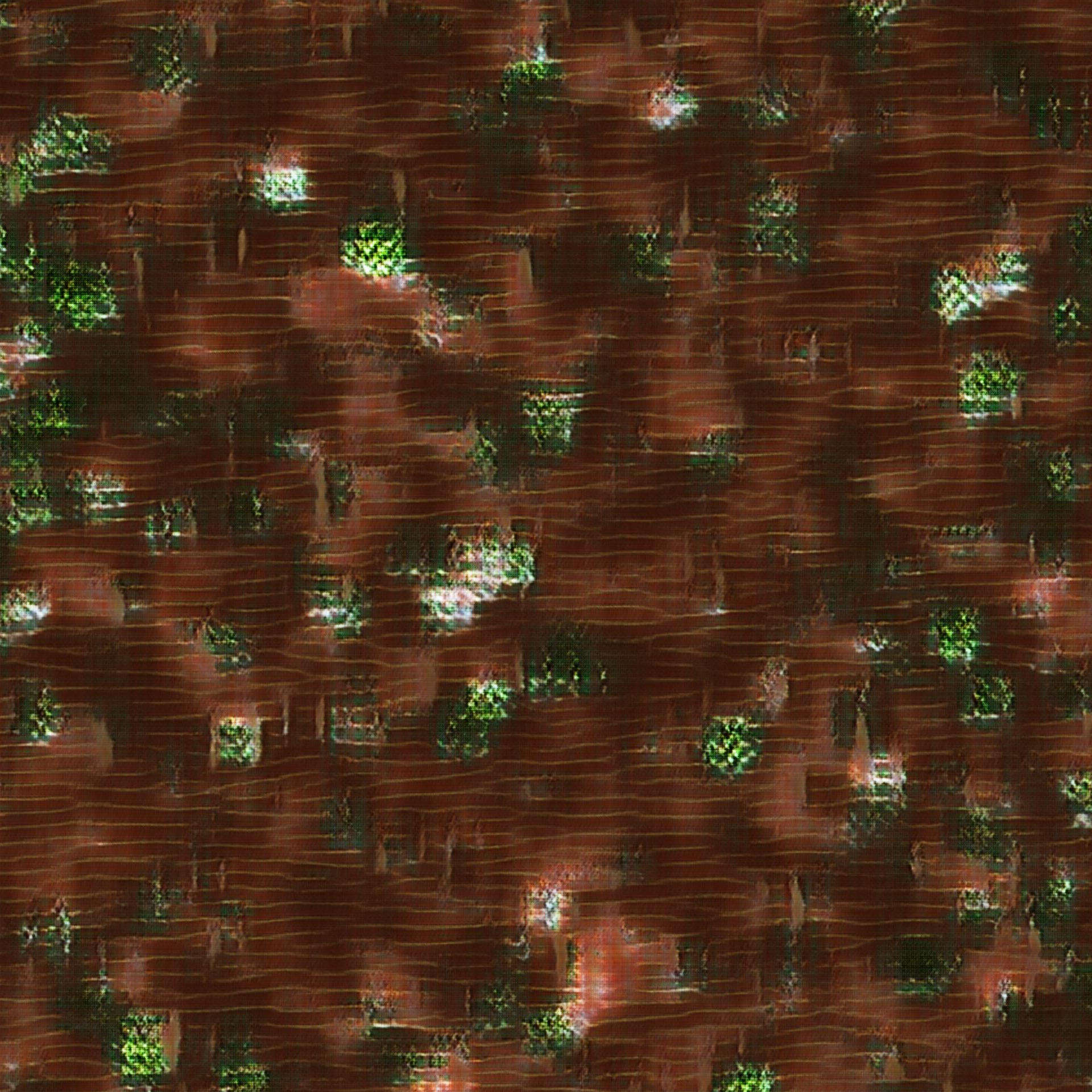} }
\subfigure[Gatys et al.]{\includegraphics[height=4.1cm]{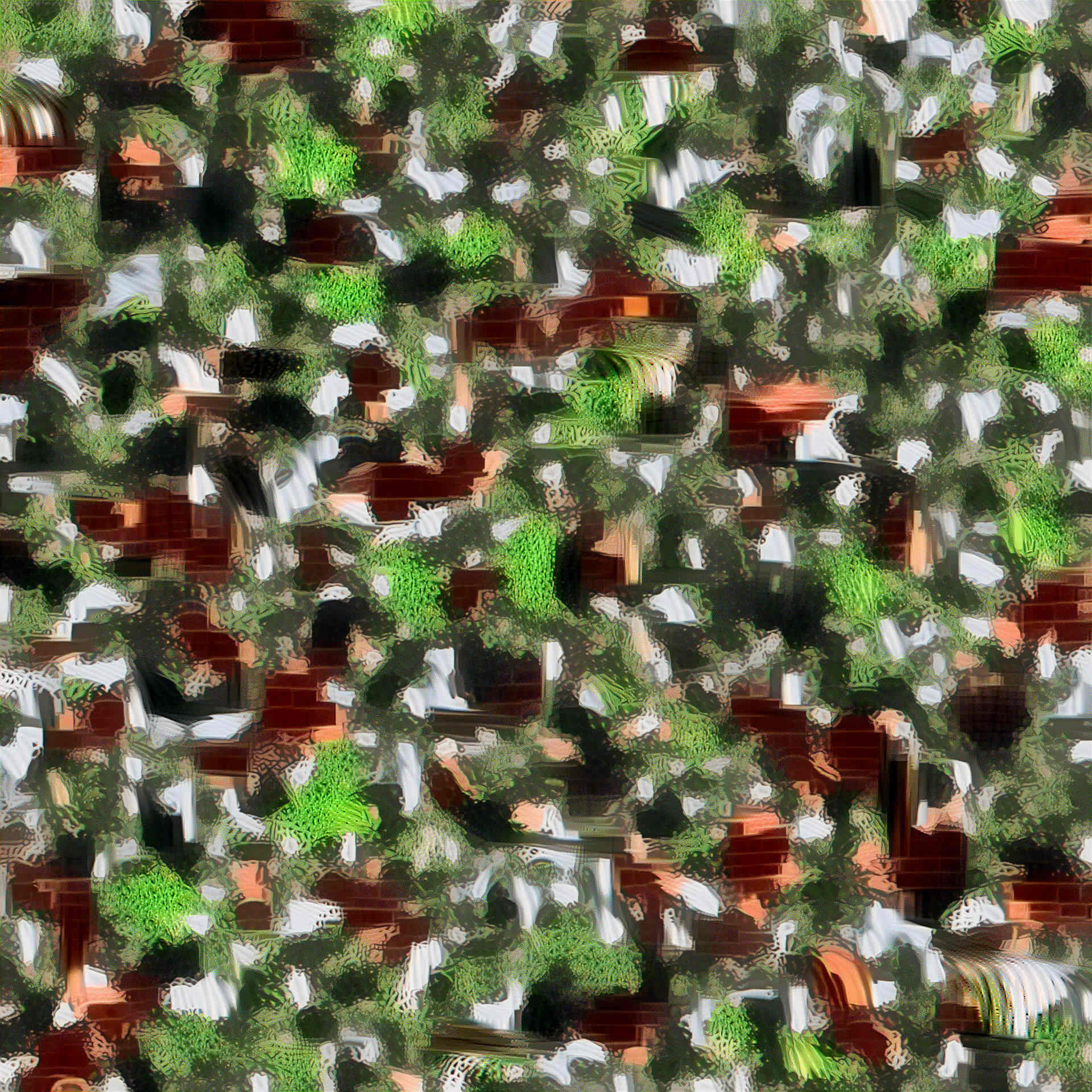} }
\vspace*{-0.4cm}
\caption{Learning from a single large photo (the Merrigum House, 3872x2592 pixels). A 1920x1920 clip-out is shown in order to have the same scale as the generated textures. PSGAN can extract multiple varied textures (bricks, grass and bushes), samples from which are shown in a 4x4 quilt, where each tile has size $\Delta=15$, for a total of 1920x1920 pixels in the generated image. Both Gatys' method and the SGAN mix the whole image instead.}\label{fig_rocinha}
\end{figure*}

Figure~\ref{fig_rocinha} shows results when trained on a single large image. PSGAN extracts  diverse bricks, grass and leaf textures.
In contrast, SGAN forces the output to be a single mixing process, rather than a multitude of different visual textures. Gatys' method also learns a single texture-like process with statistics from the whole image. \footnote{As a technical note, the whole image did not fit in memory for Gatys' method, so we trained it only on a 1920x1920 clip-out.}

\begin{figure*} 
\centering
\subfigure[\textbf{A} Sydney]{\includegraphics[height=5.5cm]{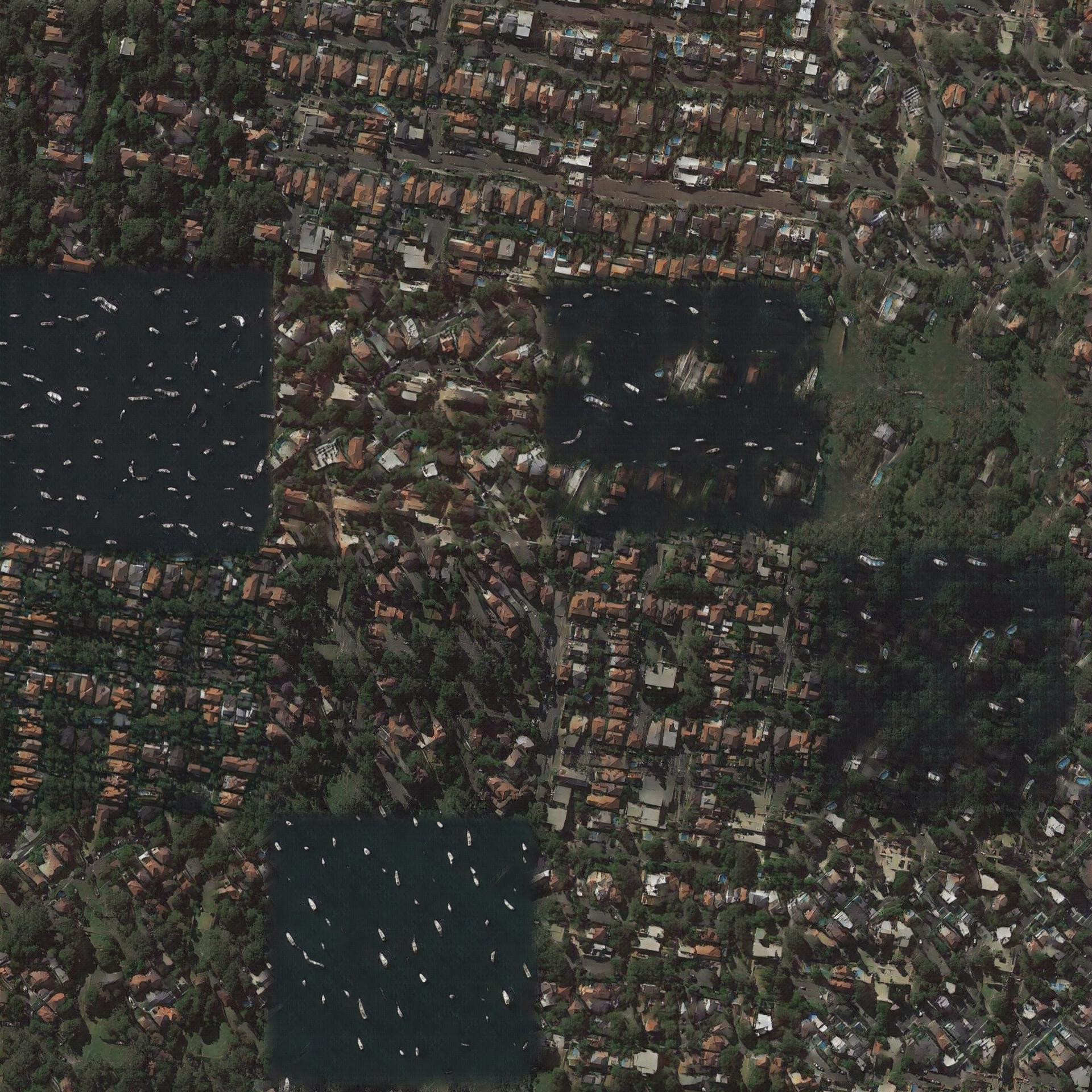} }
\subfigure[\textbf{B} DTD ``braided"]{\includegraphics[height=5.5cm]{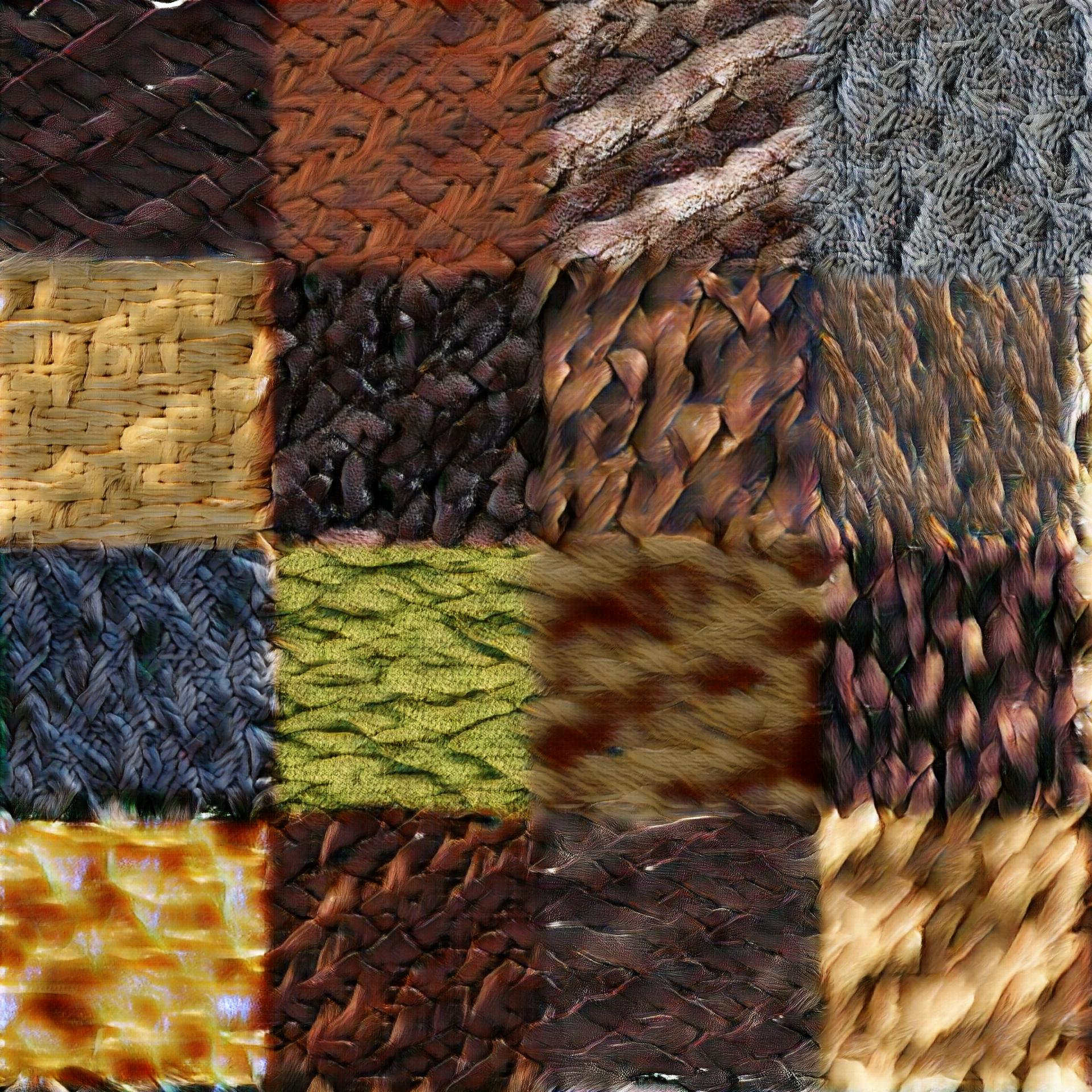} }%trim={500px 500 0 0},clip,
\subfigure[\textbf{C} DTD ``honeycomb"]{\includegraphics[height=5.5cm]{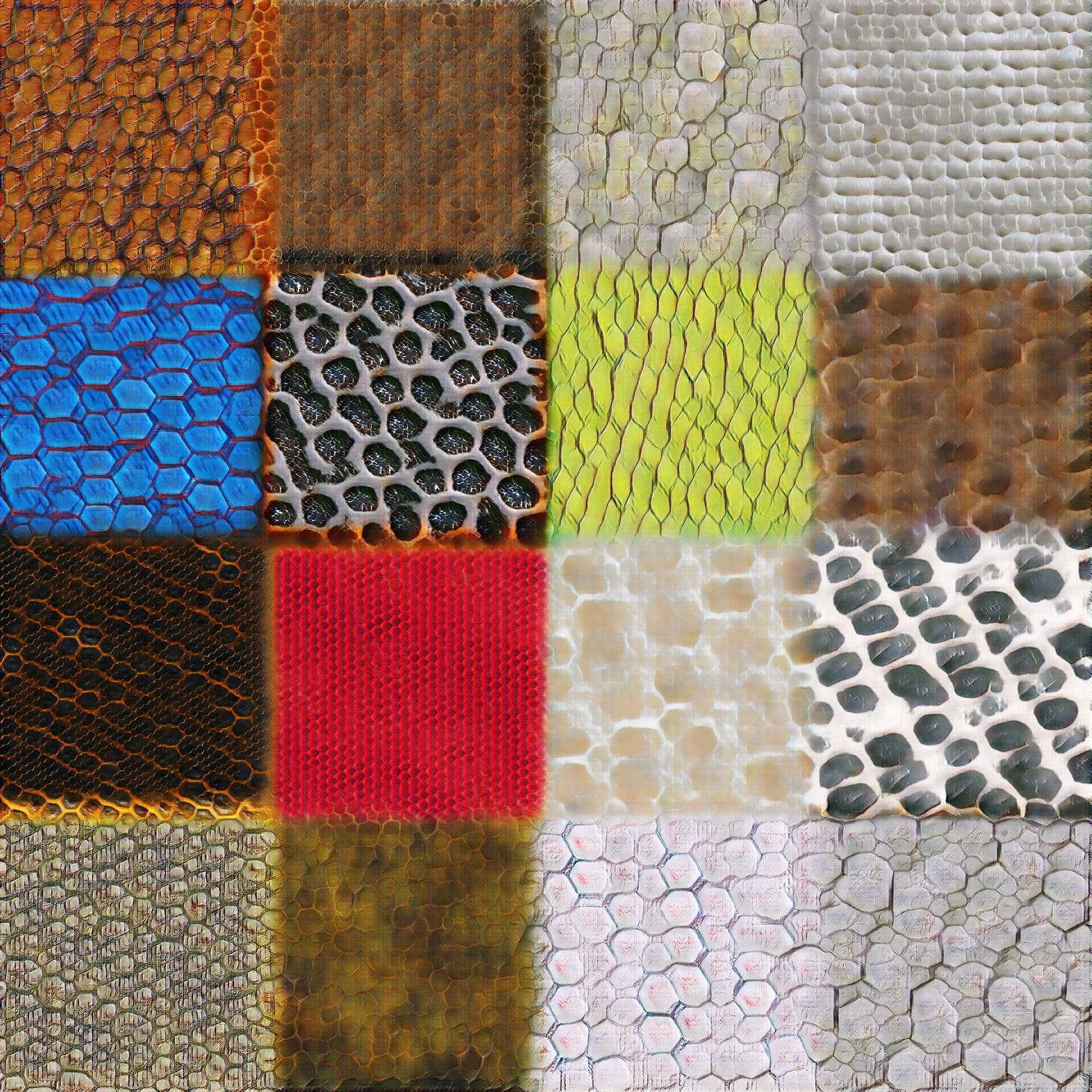}}
\vspace*{-0.4cm}
\caption{More examples of learned textures, using rich and variable input image information. \textbf{A} uses 5 satellite images (1300x700 pixels) of Sydney.  \textbf{B,C} use 120 small texture images. The outputs show 4x4 different textures on the quilt ($\Delta=15$) sampled from the model, total image size 1920x1920 pixels, best seen when maximally zoomed-in. }\label{satellite}
\end{figure*}

Figure~\ref{satellite}\textbf{A} shows texture learning from city satellite images, a challenging image domain due to fine details of the images. 
Figures~\ref{satellite}\textbf{B} and \textbf{C} show results from training on a set of multiple texture-like images from DTD.

\begin{figure}[tb]
\centering
\includegraphics[height=8.2cm]{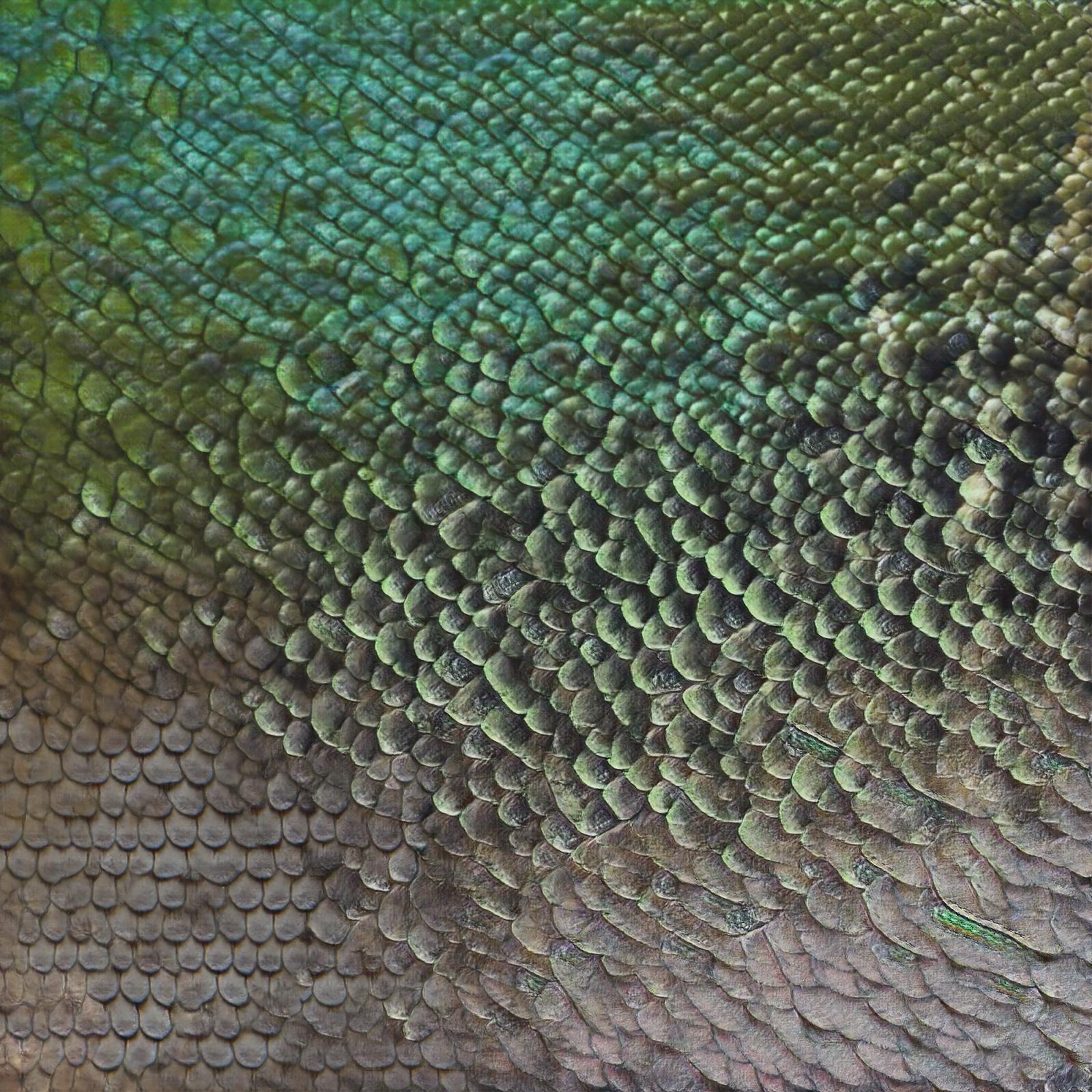} 
\vspace*{-0.3cm}
\caption{PSGAN learns a whole texture manifold from  DTD ``scaly", allowing smooth texture morphing, here illustrated in a single image of size 1600x1600 pixels. All regions of that image are plausible textures. The generator has as input a tensor $Z^g$ ($L=M=50$ spatial dimensions), calculated by bi-linear interpolation between 4 randomly sampled $\bm{z}^g$ in the 4 corners. }\label{scalemorph}
\end{figure}

In order to show that textures vary smoothly, we sample 4 different $\bm{z}^g$ values in the four corners of a target image and then interpolate bi-linearly between them to construct the $Z^g$ tensor. Figure~\ref{scalemorph} shows that all $\bm{z}^g$ values lying between the original 4 points generate proper textures as well. Hence, we speak of a learned texture manifold.

\subsubsection{Disentangling frequencies and global dimensions}

\begin{figure}[ht]
\centering
\includegraphics[height=2.4cm]{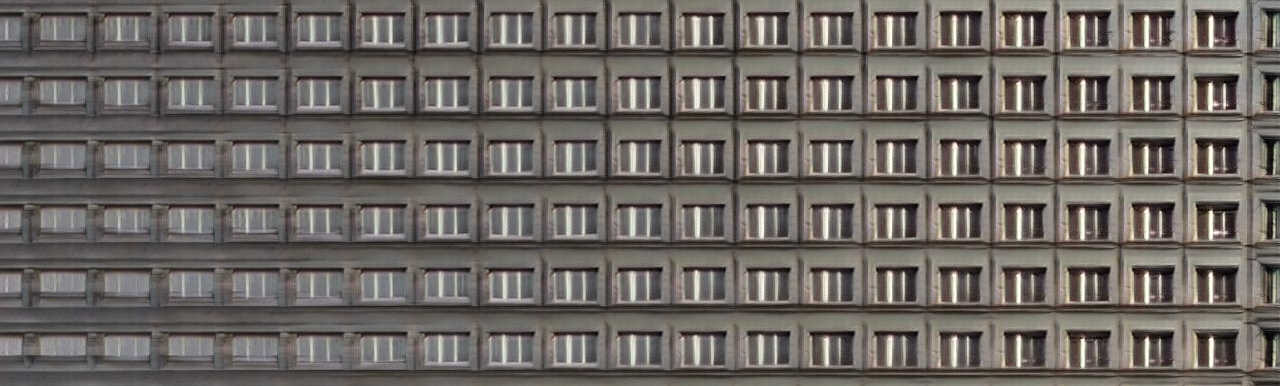} 
\vspace*{-0.4cm}
\caption{Morphing of house textures by linearly interpolating between two different textures. The disentangling properties of PSGAN allows to morph in a controlled manner: the house window periodicity stays the same, but the facade type and appearance change significantly due to the changing global dimensions.}\label{fig_disentangle2}
\end{figure}

In this section, we explore how $Z^g$ and $Z^p=\varphi(Z^g)$ -- the global and periodic dimensions -- influence the output $G([Z^l,Z^g,Z^p])$ generated from the noise tensor. 
Take a $Z^g$ array with quilt structure. 
We define as $\hat{Z}^g$ an array of the same size as $Z^g$, where all $\hat{Z}^g_{\lambda,\mu}$ are set to the same $\bm{z}^g$. We calculate two different periodic tensors, $Z^p =\varphi(Z^g)$: the first tensor with wave numbers varying as a function of the different elements of the quilt,  and the second tensor, $\hat{Z}^p = \varphi(\hat{Z^g})$, with the same wave numbers everywhere.

\begin{figure*}[ht]
\centering
\subfigure[\textbf{A} different global and periodic]{\includegraphics[height=3.8cm]{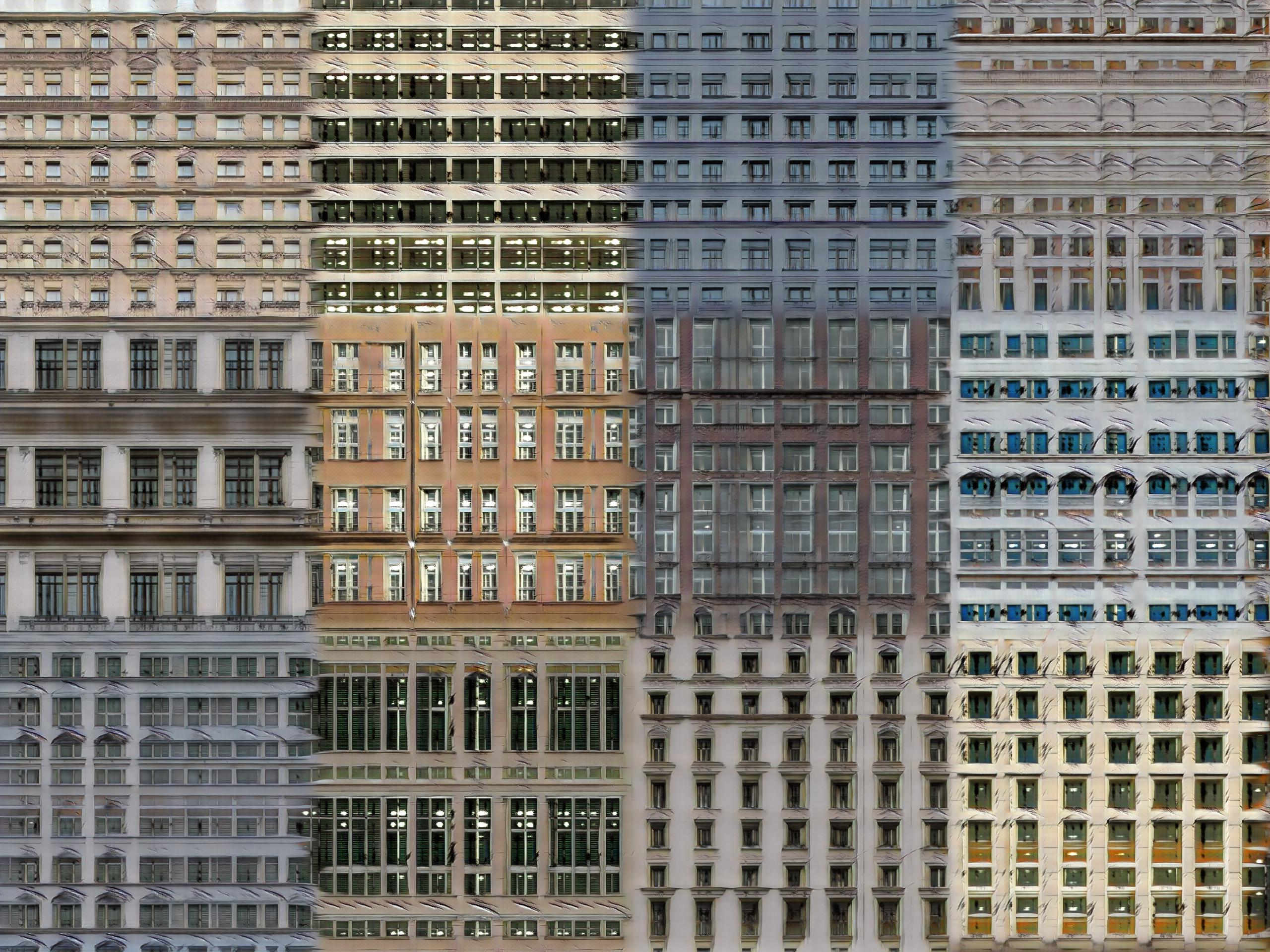} }
\subfigure[\textbf{B} same periodic dimensions]{\includegraphics[height=3.8cm]{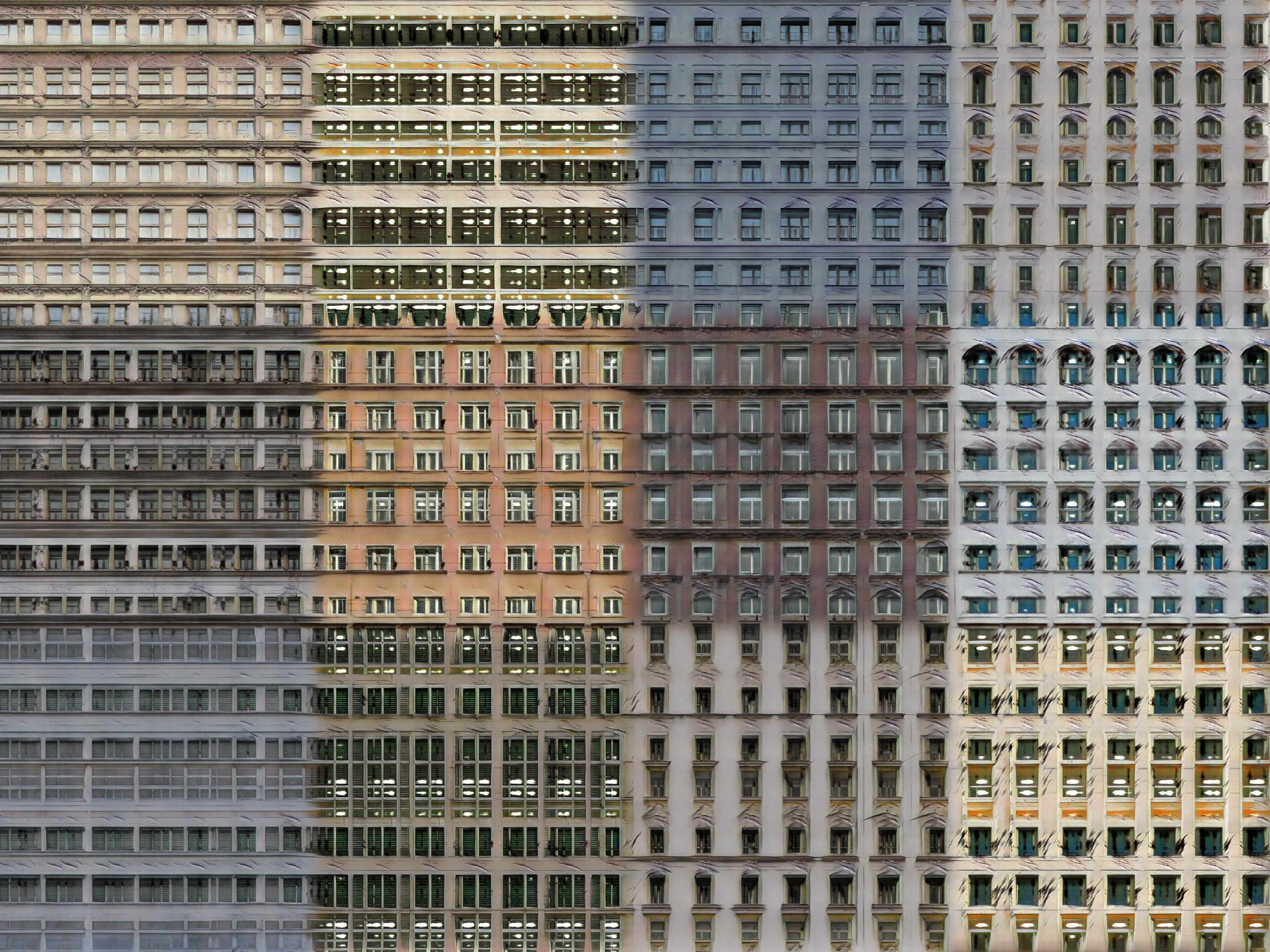}} 
\subfigure[\textbf{C} same global dimensions]{\includegraphics[height=3.8cm]{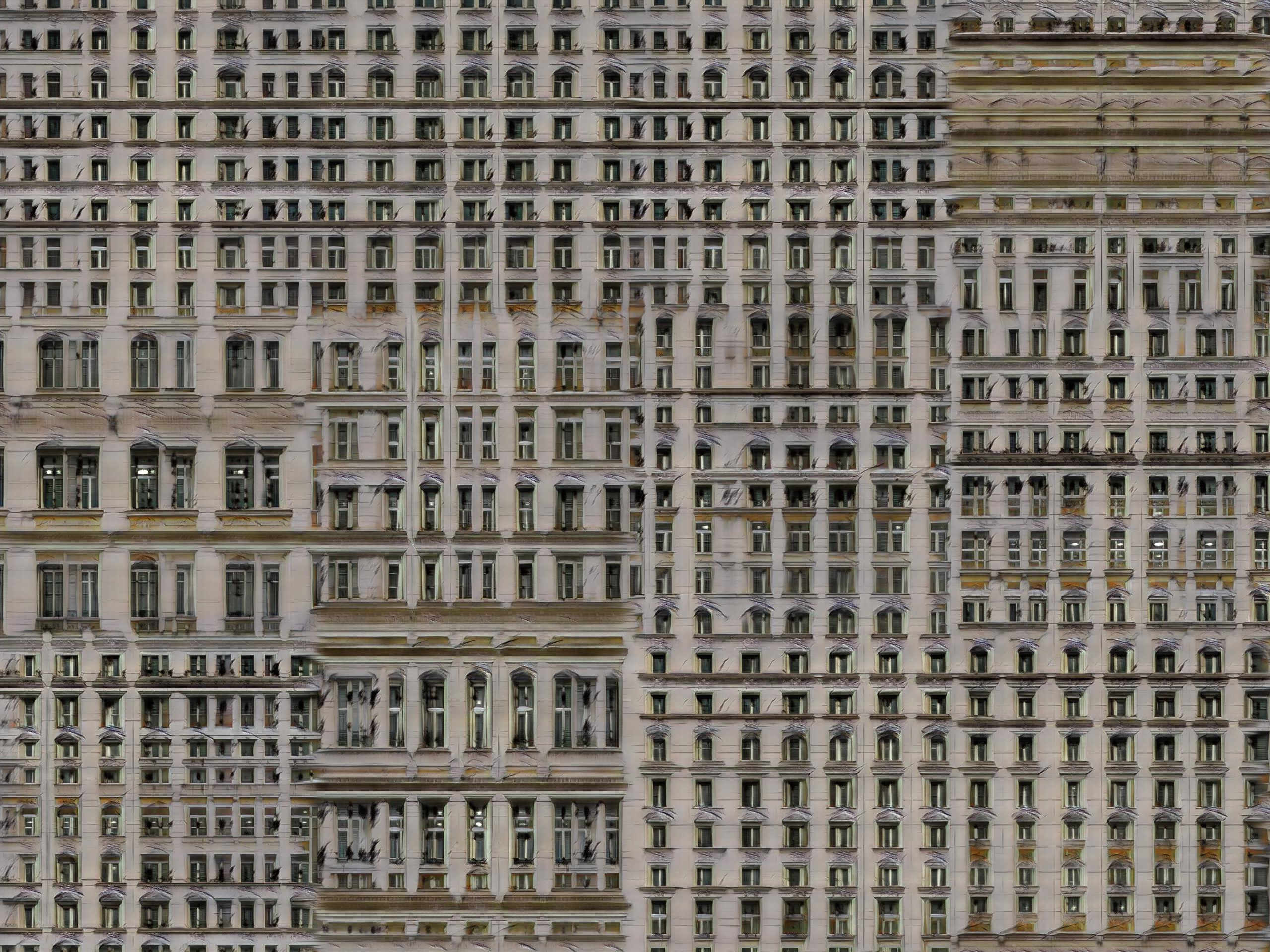} }
\caption{The influence of global and periodic dimensions of the noise tensor on texture appearance.
\textbf{A} shows the generation of the image quilt $X=G([Z^l,Z^g,Z^p])$, resulting in houses with different material and window periodicity; \textbf{B} shows $X=G([Z^l,Z^g,\hat{Z}^p]) $ -- houses with different material and color, but the same aligned periodical structure (7 windows fit horizontally in each tile); 
\textbf{C} shows $X=G([Z^l,\hat{Z}^g,Z^p])$ -- same color but different window periodicity.  The local dimensions $Z^l$ are fixed. }\label{fig_disentangle}
\end{figure*}

The PSGAN is trained with minibatches for which it holds that $Z^p=\varphi(Z^g)$, but the model is flexible and produces meaningful outputs even when setting $Z^g$ and $Z^p$ to different values.
Figure~\ref{fig_disentangle} shows that the global and periodic dimensions encode complementary aspects of the image generation process: texture identity and periodicity. 
The facades dataset has strong vertical and horizontal periodicity which is easily interpretable -- the height of floors and window placement directly depends on these frequencies.

This disentangling leads to instructive visualizations. Figure~\ref{fig_disentangle2} shows the generation from a tensor $Z^g$, which is constructed as a linear interpolation between two sampled $\bm{z}_g$ at the left and right border. However, the wave numbers of the periodic dimensions are fixed, independently of the changing global dimensions. The figure clearly shows a change in visual appearance of the texture (controlled by the global dimensions), while preserving a consistent periodic structure (fixed by the constant wave numbers). This PSGAN disentangling property is reminiscent of the way~\cite{infogan} construct categorical and continuous noise variables, which explain factors of variation such as object identity and spatial transformation.

\section{Discussion}
Texture synthesis from large unlabeled image datasets requires novel data-driven methods, going beyond older techniques that learn from single textures and rely on pre-specified statistical descriptors. Previous methods like SGAN are limited to stationary, ergodic and stochastic textures -- even if trained on many images, SGAN fuses them and outputs a single mixing process for them. 
Our experiments suggest that Gatys' method exhibits similar limitations. 
In contrast, PSGAN models non-ergodic cyclostationary processes, and can learn a whole texture manifold from sets of images, or from a single large image.

CGANs~\cite{MirzaO14} use additional label information as input to the GAN generator and discriminator, which allows for class conditional generation.
In comparison, the PSGAN also uses additional information in the generator input (the specifically designed periodic dimensions $Z^p$), but not in the discriminator. Our method remains fully unsupervised and uses only sampled noise, unlike CGANs which require specific label information.

Concerning the model architecture, the SGAN~\cite{SGAN2016} model is similar -- it can be seen as an ablated PSGAN instance with $d^g=0,d^p=0$.
This architecture allows great scalability (linear memory and runtime complexity w.r.t.\  output image pixel size) of the PSGAN when generating outputs. High resolution images can be created by splitting parts of the arrays and rendering them sequentially, thus having a constant GPU memory footprint. 
Another nice property of our architecture is the ability to stitch seamlessly output texture images and get tileable textures, potentially increasing the output image size even more.

To summarize, these are the key abilities of the PSGAN:
\begin{itemize}
\item learn textures of great variability from large images 
\item learn periodical textures 
\item learn whole manifolds of textures and smoothly blend between their elements, thus creating novel textures
\item generate images of any desired size with a fast forward pass of a convolutional neural network
\item linear scalability in memory and speed w.r.t.\ output image size.
\end{itemize}

Our method has a few limitations: convergence can be sometimes tricky, as noted for other GAN models~\cite{RadfordMC15}; like GANs, the PSGAN can suffer from ``mode dropping"  -- given a large set of textures it may learn only some of them, especially if the data varies in scale and periodicity. 
Finally, PSGANs can represent arbitrary probability distributions that extend in spatial scale to the largest periods in $Z^p$, and can generalize to periodic structures beyond that. However, images that have larger structures or more general non-periodic features are not representable: e.g.\ images with a global trend, or with a perspective projection, or aperiodic images, like Penrose tilings.

\subsection{Future work}
The PSGAN has a great potential to be adapted to further use cases.
In-painting is a possible application - our method can fill random missing image regions with fitting textures.
Texture style transfer -- painting a target image with textures -- can be done similar to the way the quilts in this paper were constructed. 
Further, explicit modeling with periodic dimensions in the PSGAN could be a great fit in other modalities, in particular time-series and audio data. Here, we'd expect the model to extract ``sound textures", which might be useful in synthesizing completely novel sounds by interpolating on the manifold.

On the theoretical side, to capture more symmetries of texture images, one could extend the $Z$ tensor even further, by adding dimensions with reflection or rotation symmetries. In terms of model stability and convergence, we'll investigate alternative GAN training criteria~\cite{MetzPPS16,wgan2017}, which may alleviate the mode dropping problem.

\section*{Acknowledgements}
We would like to thank Christian Bracher for his valuable feedback on the manuscript.

% \clearpage
\bibliography{bibi}

\begin{thebibliography}{21}
\providecommand{\natexlab}[1]{#1}
\providecommand{\url}[1]{\texttt{#1}}
\expandafter\ifx\csname urlstyle\endcsname\relax
  \providecommand{\doi}[1]{doi: #1}\else
  \providecommand{\doi}{doi: \begingroup \urlstyle{rm}\Url}\fi

\bibitem[Amidror(2015)]{amidror2015sub}
Amidror, Isaac.
\newblock Sub-nyquist artefacts and sampling moir{\'e} effects.
\newblock \emph{Royal Society open science}, 2\penalty0 (3):\penalty0 140550,
  2015.

\bibitem[Arjovsky et~al.(2017)Arjovsky, Chintala, and Bottou]{wgan2017}
Arjovsky, Martin, Chintala, Soumith, and Bottou, L\'{e}on.
\newblock {Wasserstein GAN}.
\newblock 2017.
\newblock URL \url{http://arxiv.org/abs/1701.07875}.

\bibitem[Chen et~al.(2016)Chen, Duan, Houthooft, Schulman, Sutskever, and
  Abbeel]{infogan}
Chen, Xi, Duan, Yan, Houthooft, Rein, Schulman, John, Sutskever, Ilya, and
  Abbeel, Pieter.
\newblock Infogan: Interpretable representation learning by information
  maximizing generative adversarial nets.
\newblock \emph{CoRR}, abs/1606.03657, 2016.
\newblock URL \url{http://arxiv.org/abs/1606.03657}.

\bibitem[Cimpoi et~al.(2014)Cimpoi, Maji, Kokkinos, Mohamed, and
  Vedaldi]{cimpoi14describing}
Cimpoi, M., Maji, S., Kokkinos, I., Mohamed, S., and Vedaldi, A.
\newblock Describing textures in the wild.
\newblock In \emph{Proceedings of the {IEEE} Conf. on Computer Vision and
  Pattern Recognition ({CVPR})}, 2014.

\bibitem[Dumoulin et~al.(2016)Dumoulin, Shlens, and Kudlur]{DumoulinSK16}
Dumoulin, Vincent, Shlens, Jonathon, and Kudlur, Manjunath.
\newblock A learned representation for artistic style.
\newblock \emph{CoRR}, abs/1610.07629, 2016.
\newblock URL \url{http://arxiv.org/abs/1610.07629}.

\bibitem[Efros \& Freeman(2001)Efros and Freeman]{EfrosQ}
Efros, Alexei~A. and Freeman, William~T.
\newblock Image quilting for texture synthesis and transfer.
\newblock In \emph{Proceedings of the 28th Annual Conference on Computer
  Graphics and Interactive Techniques}, SIGGRAPH, 2001.
\newblock ISBN 1-58113-374-X.
\newblock \doi{10.1145/383259.383296}.
\newblock URL \url{http://doi.acm.org/10.1145/383259.383296}.

\bibitem[Efros \& Leung(1999)Efros and Leung]{EfrosP}
Efros, Alexei~A. and Leung, Thomas~K.
\newblock Texture synthesis by non-parametric sampling.
\newblock In \emph{Proceedings of the International Conference on Computer
  Vision}, 1999.
\newblock ISBN 0-7695-0164-8.
\newblock URL \url{http://dl.acm.org/citation.cfm?id=850924.851569}.

\bibitem[Gatys et~al.(2015{\natexlab{a}})Gatys, Ecker, and Bethge]{Gatys2015b}
Gatys, Leon, Ecker, Alexander, and Bethge, Matthias.
\newblock Texture synthesis using convolutional neural networks.
\newblock In \emph{Advances in Neural Information Processing Systems 28},
  2015{\natexlab{a}}.
\newblock URL \url{http://arxiv.org/abs/1505.07376}.

\bibitem[Gatys et~al.(2015{\natexlab{b}})Gatys, Ecker, and Bethge]{GatysEB15a}
Gatys, Leon~A., Ecker, Alexander~S., and Bethge, Matthias.
\newblock A neural algorithm of artistic style.
\newblock \emph{CoRR}, abs/1508.06576, 2015{\natexlab{b}}.
\newblock URL \url{http://arxiv.org/abs/1508.06576}.

\bibitem[Georgiadis et~al.(2013)Georgiadis, Chiuso, and Soatto]{DCC2013}
Georgiadis, G., Chiuso, A., and Soatto, S.
\newblock Texture compression.
\newblock In \emph{Data Compression Conference}, March 2013.

\bibitem[Goodfellow et~al.(2014)Goodfellow, Pouget{-}Abadie, Mirza, Xu,
  Warde{-}Farley, Ozair, Courville, and Bengio]{Goodfellow14}
Goodfellow, Ian~J., Pouget{-}Abadie, Jean, Mirza, Mehdi, Xu, Bing,
  Warde{-}Farley, David, Ozair, Sherjil, Courville, Aaron~C., and Bengio,
  Yoshua.
\newblock Generative adversarial nets.
\newblock In \emph{Advances in Neural Information Processing Systems 27}, 2014.

\bibitem[Jetchev et~al.(2016)Jetchev, Bergmann, and Vollgraf]{SGAN2016}
Jetchev, Nikolay, Bergmann, Urs, and Vollgraf, Roland.
\newblock Texture synthesis with spatial generative adversarial networks.
\newblock \emph{CoRR}, abs/1611.08207, 2016.
\newblock URL \url{http://arxiv.org/abs/1611.08207}.

\bibitem[Johnson et~al.(2016)Johnson, Alahi, and
  Fei-Fei]{Johnson2016Perceptual}
Johnson, Justin, Alahi, Alexandre, and Fei-Fei, Li.
\newblock Perceptual losses for real-time style transfer and super-resolution.
\newblock In \emph{European Conference on Computer Vision}, 2016.

\bibitem[Kingma \& Ba(2014)Kingma and Ba]{KingmaB14}
Kingma, Diederik~P. and Ba, Jimmy.
\newblock Adam: {A} method for stochastic optimization.
\newblock \emph{CoRR}, abs/1412.6980, 2014.
\newblock URL \url{http://arxiv.org/abs/1412.6980}.

\bibitem[Li \& Wand(2016)Li and Wand]{LiW16b}
Li, Chuan and Wand, Michael.
\newblock Precomputed real-time texture synthesis with {M}arkovian generative
  adversarial networks.
\newblock \emph{CoRR}, abs/1604.04382, 2016.

\bibitem[Metz et~al.(2016)Metz, Poole, Pfau, and Sohl{-}Dickstein]{MetzPPS16}
Metz, Luke, Poole, Ben, Pfau, David, and Sohl{-}Dickstein, Jascha.
\newblock Unrolled generative adversarial networks.
\newblock \emph{CoRR}, abs/1611.02163, 2016.
\newblock URL \url{http://arxiv.org/abs/1611.02163}.

\bibitem[Mirza \& Osindero(2014)Mirza and Osindero]{MirzaO14}
Mirza, Mehdi and Osindero, Simon.
\newblock Conditional generative adversarial nets.
\newblock \emph{CoRR}, abs/1411.1784, 2014.
\newblock URL \url{http://arxiv.org/abs/1411.1784}.

\bibitem[Portilla \& Simoncelli(2000)Portilla and Simoncelli]{Portilla:2000}
Portilla, Javier and Simoncelli, Eero~P.
\newblock A parametric texture model based on joint statistics of complex
  wavelet coefficients.
\newblock \emph{Int. J. Comput. Vision}, 40\penalty0 (1), October 2000.
\newblock \doi{10.1023/A:1026553619983}.
\newblock URL \url{http://dx.doi.org/10.1023/A:1026553619983}.

\bibitem[Radford et~al.(2015)Radford, Metz, and Chintala]{RadfordMC15}
Radford, Alec, Metz, Luke, and Chintala, Soumith.
\newblock Unsupervised representation learning with deep convolutional
  generative adversarial networks.
\newblock \emph{CoRR}, abs/1511.06434, 2015.
\newblock URL \url{http://arxiv.org/abs/1511.06434}.

\bibitem[Radim~Tyle{\v c}ek(2013)]{Tylecek13}
Radim~Tyle{\v c}ek, Radim~{\v S}{\' a}ra.
\newblock Spatial pattern templates for recognition of objects with regular
  structure.
\newblock In \emph{Proc. GCPR}, Saarbr\"ucken, Germany, 2013.

\bibitem[Ulyanov et~al.(2016)Ulyanov, Lebedev, Vedaldi, and
  Lempitsky]{ulyanov16texture}
Ulyanov, Dmitry, Lebedev, Vadim, Vedaldi, Andrea, and Lempitsky, Victor.
\newblock Texture networks: Feed-forward synthesis of textures and stylized
  images.
\newblock In \emph{International Conference on Machine Learning}, 2016.

\end{thebibliography}
\bibliographystyle{icml2017}

\end{document}